\documentclass[12pt]{article}
\usepackage[utf8]{inputenc}
\usepackage{setspace}
\usepackage{verbatim}
\usepackage{tabularx}
\usepackage{colortbl}
\usepackage{natbib}
\usepackage{hhline}

\usepackage{multirow}

\usepackage{times} 
\usepackage[top=2.5cm, bottom=2.5cm, left=2.5cm, right=2.5cm]{geometry}
\usepackage{natbib}
\usepackage{amsmath}
\usepackage{amssymb}
\usepackage{latexsym}
\usepackage{sectsty}
\usepackage{amsfonts}
\usepackage{epsfig}
\usepackage{url}
\usepackage{subfigure}
\usepackage{lineno}
\usepackage{wrapfig}
\usepackage{comment}

\usepackage{amsmath}
\usepackage{amssymb}
\usepackage{latexsym}
\usepackage{sectsty}
\usepackage{amsfonts}
\usepackage{epsfig}
\usepackage{url}
\usepackage{subfigure}
\newcommand{\zap}[1]{}

\usepackage{color}   

\usepackage{color}   
\newcommand{\sout}[1]{ } 

\title{Can We Distinguish Machine Learning from Human Learning?}
\author{Vicki Bier, Paul B. Kantor, Gary Lupyan, Xiaojin Zhu \\ University of Wisconsin, Madison \\ \{vicki.bier@, pkantor@, lupyan@, jerryzhu@cs.\}wisc.edu }
\date{\today}

\begin{document}
\maketitle

\begin{abstract}
What makes a task relatively more or less difficult for a machine compared to a human? Much AI/ML research has focused on expanding the range of tasks that machines can do, with a focus on whether machines can beat humans. Allowing for differences in scale, we can seek interesting (anomalous) pairs of tasks T, T’.  We define interesting in this way: The “harder to learn” relation is reversed when comparing human intelligence (HI) to AI.
While humans seems to be able to understand problems by formulating rules, ML using neural networks does not rely on constructing rules.
We discuss a novel approach  where the challenge is to ``perform well under rules that have been created by human beings.'' We suggest that this  provides a rigorous and precise pathway for understanding the difference between the two kinds of learning. Specifically, we suggest a large and extensible class of learning tasks, formulated as learning under rules. With these tasks,  both the AI and HI will be studied with rigor and precision.
The immediate goal is to find interesting groundtruth rule pairs. In the long term, the goal will be  to understand, in a generalizable way, what distinguishes interesting pairs from ordinary pairs, and to define saliency behind interesting pairs. This may open new ways of thinking about AI, and provide unexpected insights into human learning.
\end{abstract}

\section{Introduction}

\begin{figure}
\centering
\includegraphics[width=14.5cm]{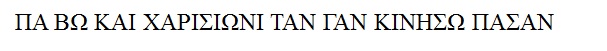}
\caption{The claim attributed to Archimedes, \cite{noauthor_quotations_nodate} }
\label{fig:Archimedes}
\end{figure}

There is enormous interest in, and confidence regarding, Machine Learning. The situation is reminiscent of Archimedes's observation about the power of a lever: ``Give me a lever long enough and a fulcrum on which to place it, and I shall move the world'' (Figure  \ref{fig:Archimedes}.) 
 Enormous computing power is used to show that, for example, a computer can teach itself to play Go, and become better than human experts (\cite{silver2017mastering}). Similarly, an algorithm has learned to play computer games, having been informed only whether it has won or lost, and the allowed set of moves (\cite{Deep_Mind}).

More generally, there is hope that machine learning
can both augment human science and be a true partner in discovery. We propose a path for exploring one aspect of that hope. Specifically, we posit that the progress of science hinges on the discovery or formulation of ``rules.'' A rule is a compact statement that supports or justifies calculations, or laboratory procedures, which, in turn, can be validated by the world. With this is mind, we have sought an abstract task of rule discovery, which does not improperly advantage either a Machine Learner (ML) or a Human Learner (HL). Specifically, can we identify {\it classes of problems} for which humans can learn the rules better than machines, and vice versa? 


\section{Statement of the Problem}\label{sec:Problem}


 We note that a ``fair'' study of how human learning aligns, or does not align with machine learning requires first that a task is being posed to humans and machines in a sufficiently comparable way. We also need to deal with the fact that even the best learning programs today may require millions of examples to be able to tell a cat from a dog, while young children seem to learn the difference much more efficiently. Even more to the point here, computers may be presented ``training examples'' of cats as a set of 2-dimensional images. Humans experience cats as objects moving continuously in time about in 3-dimensional space, and may actively interact with them.

\begin{figure}
\centering
\includegraphics[width=14.5cm]{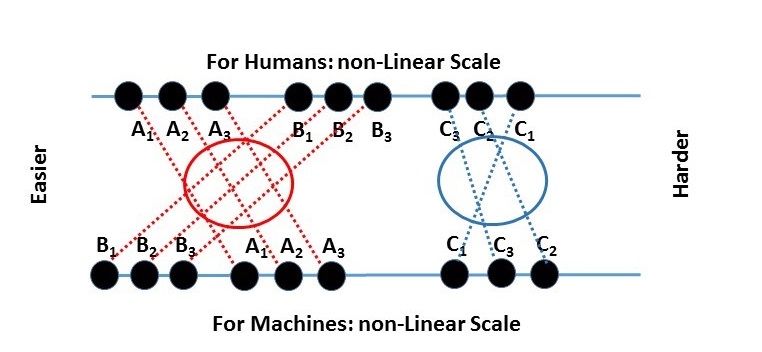}
\caption{Suppose that a number of rules in some family A have been studied, and a number of rules in family B. They are ordered in difficulty for humans along one line, and in difficulty for ML along another line. The two symbols representing the same rule are then joined. If the lines joining the symbols for two rules, A and B, must cross then we say that they form an interesting pair. See text.  }
\label{fig:crossing}
\end{figure}

We say that rule A is harder to learn than rule B if on average it takes more training episodes to learn rule A than rule B. Whether the number itself is measured in tens or millions is not the issue. What is interesting, and may provide a pathway to better understanding differences between human and machine learning, are {\it pairs} of tasks, let us call them (A,B), such that task A is harder than B for a machine but easier than B for a human, or vice versa. The idea is illustrated in Figure \ref{fig:crossing}. The relation of ``interesting pair'' is ordinal and does not depend at all on the units of measurement for either scale. 

In the example of Figure \ref{fig:crossing}, there are three classes of rules, $A,B,C$. All of the examples in Class $A$ cross all of the examples in Class $B$. When this can be found, it will provide a foundation for understanding what distinguishes the rules in those two classes. While a class, such as Class $C$ might have some internal crossings, that will not provide good information about the reasons for the crossing. 

There are several operational challenges in making this notion precise. First, for both humans and machines, the difficulty of learning a rule may depend quite strongly on the specific training sequences. The training sequences may be such as to cause one (or both) types of learning to proceed quickly. Or it may make learning difficult (that is, slower) for either, or both types of learners. Thus research must either be able to identify these order effects, or must average over a sufficiently large set of training sequences, for both ML and HL.  Since a computer can be told to forget everything it has experienced, this is relatively easy for the ML arm of the research. For the HL arm, by contrast, different participants must be employed for each training sequence, to provide a comparable {\it tabula rasa.}

A second concern is intrinsic randomness. Almost all contemporary ML approaches involve some stochasticity. We ought to average over multiple runs.
Of course, there are individual human learner differences as well, and we again must average over them. 
This suggests that such research may require hundreds or more human participants. Fortunately, such studies have become possible and accepted using Mechanical Turk techniques. 
To decide which type of learner is better at a task,
a non-parametric test such as Wilcoxon Rank test can be applied.

There is a third concern. We are interested not only in the relative learnability  of specific classes of rules, but what is perhaps even more important, to explore whether transfer from one learning task to another is the same or different for HL and ML. In order to explore this, we will have to develop concrete measures of transferablity. Conceptually, the problem is the same as the one represented in Figure \ref{fig:crossing}. However, in this case the black dots will represent transfer pairs. Thus the symbol $A_1$ would represent the amount of transfer from one specific rule, say $R_1$, to another, $R_2$.  

\section{Background to the Problem}\label{sect:Background}

Every aspect of this task has an enormous and relevant literature. In this brief summary we can point to only a few of the publications that provide a framework for the task posed here.

\subsection{Human Learning}\label{subsec:HumLearn}
Many of the classic domains that were once viewed  as pinnacles of human intelligence (chess, logical reasoning) have been conquered by relatively simple algorithms. Conversely, tasks that are easy for humans and other animals —  such as flexible locomotion and rapid and robust visual categorization — are all at the cutting edge of modern artificial intelligence research. This is known as Moravec’s paradox.

Consider, for example, that a simple electronic calculator can do arithmetic orders of magnitude faster than a brain made up of a 100 billion neurons.  It is not just that a calculator is so much faster than a person. When applying simple algorithms, people make mistakes that computers never make. For example, a large minority of people who are able to correctly define what makes a number even or odd, but nevertheless systematically misclassify numbers like 798 as odd. People who  know that a triangle is a three-sided polygon nevertheless claim that equilateral triangles are “better” triangles than scalene triangles, and frequently misclassify the latter type of triangle as not a triangle at all ( \cite{lupyan_difficulties_2013}.
)

Cognitive science and neuroscience help us understand what is going on here. What makes computers fast and accurate is their ability to perform simple computations with high precision. Biological computation is in comparison much slower, and — critically — much noisier. This means that long serial computations (even as ‘simple’ as binary addition) cannot be achieved with high precision. Biological neural networks compensate through the use of massively parallel and distributed  computation, an observation presciently made by von Neumann more than half a century ago (reprinted as \cite{von_neumann_computer_2012}). 

This parallel and distributed architecture is ill-suited for carrying out the kinds of computations (arithmetic, logic), that are trivial for electronic circuits. The reason people mis-classify 798 as an odd number is not that they are inattentive or careless. Rather, applying abstract rules — even very simple ones like `MOD 2,' requires representing only the parts of the input that matter — here, whether a number is evenly divisible by 2. This requires projecting the original representation (which contains information about the number, its magnitude, the color of the font that comprises it, its location in space, etc.) to a space with a discrete decision boundary (even/odd). In this process, a number like 798 is closer to the odd/even boundary than a “more even” number like 400; occasionally, 798 ends up on the wrong side, as reported by 
\cite{lupyan_paradox_2015, lupyan_difficulties_2013}.

The same similarity-based processing that makes it difficult for people to apply an abstract rule quickly and robustly is ideal for learning similarity-based representations and discovering (even very subtle) covariance structures present in the input (\cite{rogers_semantic_2004,rumelhart_parallel_1986}). Even six-month-old infants can learn what cats have in common that distinguishes them from dogs
(\cite{quinn_perceptual_1996}
). It is not a coincidence that research 
attempts to build categorization algorithms that approximate human categorization began to succeed only when the underlying architecture moved from rule-based ‘expert’ systems to distributed architectures (such as artificial neural networks) that rely on gradually learning from multiple examples. It is also not a coincidence that we do not use these architectures for doing arithmetic or logic.

\begin{figure}
\centering
\includegraphics[width=14.5cm]{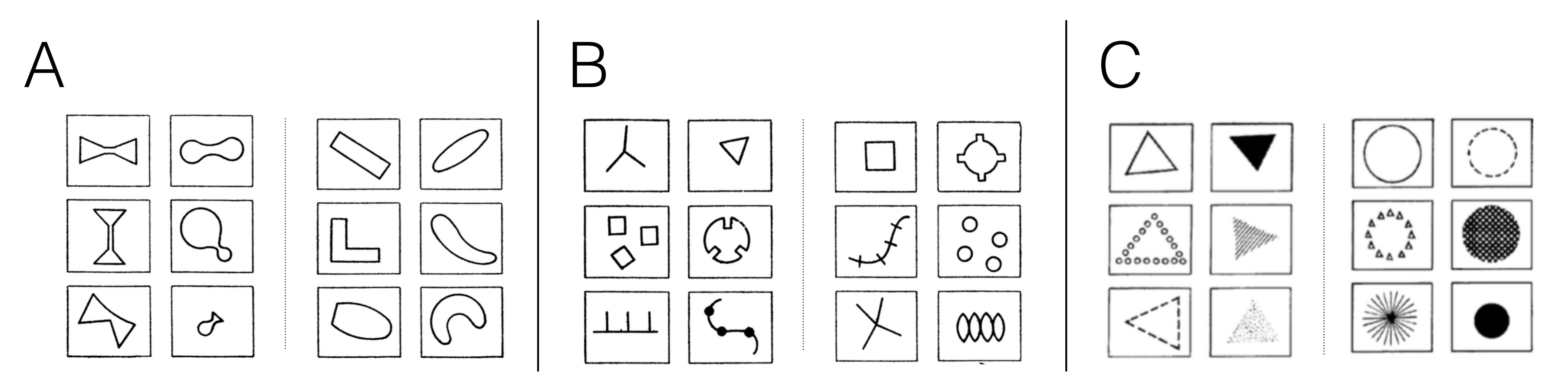}
\caption{Three Bongard Problems}
\label{fig:bongard}
\end{figure}

Despite being slow and error-prone when applying even simple rules, people are capable of remarkable acts of abstraction in tasks requiring \emph{extracting} rules from examples. A classic example of this problem domain are Bongard problems (\cite{bongard_pattern_1967}), three of which are shown in Figure  \ref{fig:bongard}. In each problem, people are presented  with 12 shapes and must formulate a rule that distinguishes the shapes on the left from the shapes on the right. The “rule” formulating the distinction in Figure \ref{fig:bongard}A is that all shapes on the left and none on the right have a narrowing in the middle. Despite the geometric simplicity of the distinction (which can be obtained through computing the polarity of the second derivative), this problem is relatively difficult for people — only 36\% of the subjects succeed. The problem shown in Figure \ref{fig:bongard}B  is easier — 76\% correctly induce that the rule is threes vs. fours. Compared to the previous problem, this one seems to require much more abstraction: the solver needs to abstract over the different instantiations of `threeness': three line segments; a triangle; three polygons; three notches; etc.  The easiest of all is the problem shown in Figure \ref{fig:bongard}C, which was solved by over 95\% of our participants. The rule — triangles vs. circles — may seem trivial, but is exceedingly difficult to derive if one does not already know about “circles” and “triangles”--the “circle” images that are shown have no features in common with one another. This challenge was first pointed out by Mikhail Bongard, the original creator of the problems (\cite{bongard_pattern_1967},  see also \cite{linhares_glimpse_2000}). What enables people to solve this problem so easily is that they come to the task having previously learned a set of higher-level categories (e.g., circle, triangle, three). They are then able to flexibly deploy them as hypotheses in a top-down way (\cite{lupyan_words_2015}). Many of these units may have been learned in the course of learning one’s native language (\cite{majid_differential_2018, lupyan_words_2015}). It is not a coincidence that the relative ease of the problems in Figures \ref{fig:bongard}B-C compared to Figure \ref{fig:bongard}A is strongly correlated with knowing words like “three,” “four”, “triangle,” and “circle”. If this analysis is correct, then the key to closing the gap between the human and machine extraction of rules may lie in understanding what units people use when extracting rules from examples, and what makes a unit especially useful.

\subsection{Machine Learning}\label{subsec:MachLearn}

When choosing a representative machine learning paradigm for a study such as is contemplated here, the main consideration is to closely match its human learner counterpart.  As such, we propose to study reinforcement learning agents: like humans, they play with possible actions, receive rewards or penalty, and update their policies.   In addition, reinforcement learning is well-studied in machine learning and has previously achieved impressive abilities in playing games like Go (\cite{silver2017mastering}). 

To specify the machine learning task, we define the rule game learning problem in terms of a Markov Decision Process $M=(S, A, T, R, \gamma)$:
\begin{itemize}

\item The state space $S$ represents the game board as well as historical plays, 
both successful and unsuccessful.

\item The action space $A$ represents possible moves.

\item The state transition probability $T$ specifies how the game is updated to new state $s'$ upon playing move $a$ in state $s$: $P(s' \mid s,a) := T(s', s, a)$.

\item The reward function $R(s,a)$ is 1 if the move $a$ is accepted, $-1$ if not.  The cumulative reward therefore penalizes long sequences of wrong moves.

\item $\gamma$ is a standard discounting factor in reinforcement learning to combine long-term rewards.
\end{itemize}

It is important to point out a subtlety in evaluation.
Since our overall task is ``rule learning,'' a natural goal might be for the machine to exactly identify the rule.  While this is certainly reasonable, it is nonetheless restrictive for several reasons. 
First, most reinforcement learning agents represent their policies (why they choose an action under a given circumstance, or state) implicitly: as a Q-table, or value function approximation, or a policy neural network.  It is difficult to extract the ``rule'' in a human readable form. 
Second, even if we could do that, we would then need a quality measure to compare the machine's rule vs. the ground-truth rule (this applies to human learners, too).  This can be difficult to do.
Finally, it is possible that the human learners say one thing (their purported rule) but do another (playing the game in a way that does not match their purported rule), which creates complications in comparing to machines.

Therefore, rather than attempting to extract rule knowledge from the machine, we will measure machine performance by how well machines can learn and play games with different rules.  More precisely, we will consider discounted reward, average per-round success rate, and speed at which these measures asymptote.
Note that these measures can be equally applied to human learners, which we plan to do (except that for human learners, we will also ask them to state their purported learned rule at the end).  Eventually, the relative difficulty of two rules to a reinforcement learning agent can be measured by the agent's learning curves for the two rules: what levels of performance the curves reach, and how fast they get there.

\subsection{Translation to Real Problems}\label{subsec:Translation}

To understand the importance of rule learning in applications of artificial intelligence, it is useful to present a bit of history. The earliest applications of artificial intelligence were typically rule-based—e.g., for medical diagnosis, with a very  early example being \cite{shortliffe1975computer} . 
For  a more nearly modern view  see \cite{lim_et_al}), or electronic checklists to support automated diagnosis of equipment problems (\cite{Fung_1989})—meaning that the rules were explicitly programmed by humans, rather than being learned endogenously by machines.
These applications, while useful at improving efficiency and reducing tedium, could never significantly outperform the best human experts.  They could allow novices to achieve near-expert performance, and improve the consistency and accuracy of human experts, but since they were dependent on “handcrafted knowledge,” they were inherently limited by human capabilities.  In other words, rule-based AI could be be “faster and less error-prone,” and have “a higher degree of precision,” but represent at best an incremental improvement over human capability.  
More recently, there has been an explosion of AI capabilities, due to the adoption of machine learning and statistical pattern recognition.  This has resulted in truly spectacular achievements, such as the development of a computer program (Alpha Go) that has far outstripped the world’s best Go master in a remarkably short period of time.  Thus, it would appear that we are now in an era where artificial intelligence has surpassed human intelligence, 
although the ``rules
learned'' 
by machines are embedded in a truly opaque forest of linking parameters.

However, artificial intelligence still has some dramatic limitations.  In particular, it works best in highly constrained environments (e.g., games such as Go or chess), where it is clear which types of “moves” or rules are permitted (even if the software needs to learn the rules on its own by observation  (\cite{Deep_Mind}).  In less structured environments, it often performs poorly (or at least non-intuitively), making “mistakes” or misinterpretations that in some cases would have been obviously (or hilariously) wrong to even the most naive human subject; see for example \cite{goodfellow_attacking_2017}, \cite{krakovna_specification_2018}. 
      
As a result, many applications of AI are still quite “small.” Even though deep neural networks are now capable of recognizing objects in a complex visual field, they are still of limited reliability in the real world.  Thus, for example, in development of software for identifying skin cancer, “if an image had a ruler in it, the algorithm was more likely to call a tumor malignant” (\cite{Patel_Daily_Beast}). 
Visual recognition can automate the review of vast quantities of visual data; even if the process is error-prone, it can still be useful by flagging potential targets for human review.   
 
However, the task is limited by the binary (or near-binary) nature of the response variable.  AI is typically not used to recognize every item in a complex visual field, only to flag those that meet specified criteria.  When the classification process is more open-ended, image recognition can still yield surprising errors.  In some cases, these errors are not too bad, and might also be made by humans (e.g., incorrectly classifying a comforter as a pillow, or a dog as a cat or wolf). In other cases, however, the errors are more serious; for example, erroneously classifying a turtle as a rifle (or the reverse) could have significant adverse consequences (\cite{molnar}, \cite{turtle_example}). Moreover, even for a single item, it is not difficult to fool an algorithm; an interactive example is given by \cite{papernot_how_2019}.

Therefore, humans are still needed for higher-level tasks—e.g., making decisions of what to do about objects after they have been recognized by machine learning.  This is especially true in situations where the decisions have high stakes (e.g., deciding on a medical treatment, rather than a chat bot deciding on which product to recommend to a customer).  Thus, suitability for machine learning (\cite{suitability_machine_learning}) is judged to be low if a task requires “complex, abstract reasoning,” while computers are more suitable for routine repetitive tasks, where efficiency is prized and the cost of errors may be low.  Machine learning can also be vulnerable to adversarial attacks such as fraud (\cite{levin_jerry}).  

The proposed research—identifying which types of rules (or changes in rules) are more easily learned by humans, and which are more easily learned by computers—could pave the way for more complete human-assisted AI (or AI-assisted human decision making), in which computers can take over more complex functions, but in a gradual manner, consistent with a thorough understanding of their capabilities.  As stated by \cite{polson_aiq:_2018}, AI can yield “different and better jobs, new conveniences, freedom from drudgery, safer workplaces, better health care, fewer language barriers, new tools for learning and decision-making that will help us all be smarter, better people.”

\section{Game as a Learning Task}\label{sect:model}
\begin{figure}
\centering
\includegraphics[width=14.5cm]{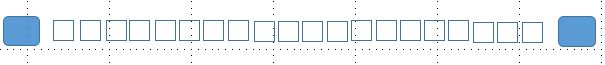}
\caption{An example display might have some number, in this case, 20, of positions along a line. There are ``buckets'' at either end of the line. A ``rule'' specifies the order in which objects are to be moved from the display. It further specifies, when an object may be moved,  the  bucket into which the object, at that  move, is to be placed.  }
\label{fig:empty}
\end{figure}

\begin{figure}
\centering
\includegraphics[width=14.5cm]{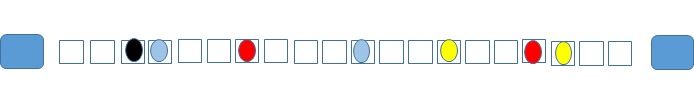}
\caption{This is an example initial configuration in which colored objects are placed in some of the positions. The initial configuration for each episode of play is to be generated randomly. }
\label{fig:someobjects}
\end{figure}

As an illustration, 
one might consider
 a learning task in which colored blocks are placed in any one of $L=20$ positions along a line, as in Figure \ref{fig:empty}. A {\it move} is to take a block and place it in a bucket at one end of the line or the other. Although there are 20 places in the lines, on any given {\it episode} of play, somewhere between five and ten colored blocks are placed randomly in some of the positions, with no more than one block in a position, as in Figure \ref{fig:someobjects}. 

One player, Alice, formulates a rule (see some examples in Exhibit 1)  and observes while a second player, Bob, tries to play the game. An episode ends in success when all of the blocks have been removed in accordance with the rule, and each has been placed in a bucket allowed by the rule.

To be concrete,  some possible rules are shown in Exhibit 1.

\noindent\fbox{
    \parbox{\textwidth}{
   \begin{center}Exhibit 1 \end{center}
\begin{enumerate}\label{list:rules}
   \item Remove items from to left to right,  placing each object in any bucket.
    \item Remove items from  left to right  placing each in the nearest bucket.
    \item Remove all blue blocks, into the   left bucket, and all red blocks into the right bucket, and all other blocks can go in either bucket
    \item Remove blocks, from the outside in, starting at the left end. Place each in the farthest bucket
    \item Place any block in any bucket except that if there is a red block in the seventh position, reading  from the  left, it must be the third item removed, and must go into the right   bucket.
    \item Place a first block in either bucket, and thereafter remove blocks in any order, placing them alternately into left and right buckets
\end{enumerate}
 }%
}

\section{How Many Rules are There? }\label{sect:size}

The size of the rule space is enormous. For example, the rule ``remove objects, from left to right, and place them all on the left'' is one of $L!$ possible orders. In addition, if there are $C$ colors, and allowing for interaction between position and color there are $2^{CL}$ possible rules, for each order.  So, for this particular instance, with, say, only three colors [our example actually has four], $C=3$,  and $L=20$ there are $20!2^{60}=2.8\times 10^{36}$ possible rules. 

In theory, this should be an advantage for machine learning, since a computer could search a much larger fraction of the space of possible rules than any human could.  In practice, however, humans may have good intuition for certain types of rules, if based on preexisting concepts (e.g., from natural language). 

\section{What is a Simple Rule?}\label{sect:rule_size}



Because the problem that we pose (about rule learning) could easily be said to encompass all of knowledge, both human and otherwise, it may be helpful to compress the space of examples to be considered.  This may necessarily be somewhat arbitrary. There are several ways to limit the space of possible rules. One is to somehow limit the expressive power of the language used to encode the rules. If that is the approach, there are still many design decisions to be made. Since a rule will not be ``interesting'' unless it can be created and learned by humans, the gold standard for these decisions is going to be what is learned from human learners participating in the proposed ``game'' (see discussion in Section \ref{sect:model} above). 

For example, one could say that the rule should be ``not too big,'' and one way to measure size could be by the number of bits in the rule, together with the number of bits in the code book (\cite{rissanen1989stochastic}). However, it is easy to create examples (as we illustrate below) that can be described in a relatively small string, but require enormous exploration to be determined. 

The situation becomes even more difficult if we replace the notion of ``learning the rule'' with some variant of being Probabilistically Approximately Correct. In this case, rules that encompass rare exceptions may be considered well learned even by learners who do not learn those exceptions. In addition, constraints might be proposed about what kinds of information about the display, and about previous moves, can enter into the rules to be considered. 



A rule will be an unambiguous statement that makes it possible, at any point during an episode, to determine whether a block can be moved at that point, and whether it has been moved to the correct bucket. 
One way of limiting the set of allowed rules is in the way that the rule for what is ``allowed'' not be permitted to depend on unsuccessful prior attempts. This would eliminate rules such as ``you must try two moves that{\it are not } allowed, before you will be permitted to make a move that {\it is} allowed.

Even when some types of rules are excluded, there are still many possible rules, as discussed in Section \ref{sect:size}. As an extreme  example, the rule could contain a precise description of one possible starting position, with blocks of colors $c_1,c_2,\dots,c_k$ occupying positions $p_1,p_2,\dots,p_k$; let us call this {\bf configuration} $\mathbf{D_1}$. A rule, let us call it $\mathbf{R1}$, might say ``if $\mathbf{D_1}$ is presented, place everything on the right; otherwise, place everything anywhere.'' While this rule appears simple, discovering it would require an enormous search. Even if given the hint that ``there is only one starting configuration for which you are not free to do whatever you like,'' the expected time to discover this particular configuration would be half the number of possible configurations.\footnote{With bad luck, one might even {\it accidentally} do the allowed thing for this configuration, and would find that out only after trying another solution for every configuration, until one is told that a move is not allowed.}
Thus it seems reasonable to eliminate rules with such a strong dependence on the initial configuration. 
For this particular game, with $K$ positions filled, there are $2^K \binom{L}{K}$ possible initial configurations.  If there were two special configurations, the ``learning process'' would take 50\% longer, in expectation. 

Of course, rules of that form could also  be disallowed.   

Another promising approach is 
to develop a specific language for the rules, and then place a limit on the number of terms from that language that can be present in an allowed rule. For example, one could permit a rule to use information about both position and color, provided that the rule can be expressed in a ``limited number of bytes.''
Of course, such a constraint would depend not only on the rule itself, but also on the cleverness of the team specifying the code book and the notation, so it may be difficult to establish whether a given rule ``can be expressed in less than N bytes.'' 

A second issue is to limit the kind of ``scratch tape'' or ``auxiliary registers'' that may be used in the process of learning a rule; for example, whether the machine-learning algorithm is allowed to track only the history of successful moves in a given episode, or also any unsuccessful move attempts.\footnote{Of course, both human and machine learners will have to remember previous episodes, in order to find a rule.} 
Here, the comparison between machine and human learners is complicated by the fact that humans will remember (some fraction of) both successful and unsuccessful moves, but may remember them imperfectly, or even erroneously.

Any particular framework for machine learning, along the lines of the above, will of course limit the aspects of history that can be used in rule learning. For example, the present move for an object of a given color
could be restricted to depend only on the {\it most recent} correct move of an object having the same color.  Thus, ``blues must be dropped alternately left and right, when they are dropped'' would be allowed under this type of rule, but ``reds must be dropped cycling around and counting the Fibonacci numbers for the positions (modulo 4)'' would not be allowed.

With regard to the treatment of past history in rule learning, there are a few key observations.  
First, one may find that  human subjects are capable of both generating and learning rules requiring a more complex treatment of history than we might initially assume. Whether this happens naturally, or can be elicited with suitable instruction, is an open question. 

Second, we do not yet know 
how such a   
study will apply to real-world rule induction situations. 
Such rules might apply to tactical issues, such as  diagnosing the problem with a portable generator. On the other hand, rules may also be sought for strategic issues, such as identifying methods used by an adversary such as a fraudster. One must, for this kind of translation of research results, explore  what kinds of rules have been proposed in the literature on these issues. While the rules will almost surely be related to the adversary's historical behavior, they will probably not contain complex mathematical concepts. 

Finally, scientific discovery is also a kind of ``rule-learning,'' where the rules are the Laws of Nature. For example, the historian of science Peter Galison (\cite{galison1997image}) has given primacy not to theories (as in the work of Thomas Kuhn (\cite{kuhn1962structure})\footnote{Note that at least one of the present authors holds that ``Kuhn is not a Kuhnian'', as explained in the post-script to the second version of his influential book \cite{kuhn_structure_1970}}, but to experiment and observation.  

It is well known that for physics, at least, a first order model (with respect to time) is clearly not adequate. For example, the distance that an object falls depends quadratically on the time. If we imagine steps in the game as a time variable, such a quadratic dependence can not be imposed if the rule must depend only on the most recent previous event. For gravity, the new increment, in this case, distance, is not given by a static rule, but must change after each new increment.  

There are  examples of physical rules that involve only the relation between a velocity and the current state of the system (and not on some ``wall clock,'' see for example \cite{carter_2003}).  As these examples suggest, any restrictions on the types of rules that are allowed in research on rule learning will somehow  limit the types of situations to which the findings of that learning can be applied.

\section{Discussion}\label{sec:discussion}

What might be gained by the investigations sketched here?  There are potential implications for both psychology and computer science. If one can find a ``comprehensible'' (to humans) distinction between rule pairs that are ``interesting'' and those that are ``uninteresting,'' that will suggest new lines of research.

\begin{itemize}
    \item For psychology: can we train people to do better on the  classes which are, by comparison, ML-easy and HL-hard?
    \item For computer science: can one  extend  learning methods to make some of the classes that are ML-hard become ``easier?''
    \item For application of this research to real-world problems, this research may lead to better harmonization of human and machine capabilities to  jointly solve complex problems in a manner consistent with their capabilities. 
  \end{itemize}{}  
    In particular, expanding on the third point,  deeper understanding of the differences between human and machine learning might make it possible to ``triage'' problems that lack known rules of procedure, and direct such problems to humans or machines, according to which has a better chance of inferring or inducing the correct rule in time for the solution to be useful.  While the successes of Machine Learning are impressive, their consumption of time and energy is a significant factor in potential application (\cite{Energy_garcia-martin_estimation_2019}). Machine Learning has shown  substantial advances on problems  for which  deterministic ``oracles'' exist (such as video games or board games). In these cases the learner is told some part of the rules (such as what moves are allowed), and the remainder is provided by an oracle. Problems of image classification appear deterministic to the machine learner, but of course the human labeling of ``ground truth'' almost certainly contains errors. 
    
    The field of generalized language understanding remains very challenging, and the largest ongoing project (\cite{lenat1995cyc}), seemed to pursue an ever-retreating horizon.  The more recent refocus on specific (multiple choice) tasks, seems to promise a path for solution to (at present) eighth grade New York State Regents examination in science. However, this success is apparently limited to chains of reasoning about synonyms and relations, and cannot (yet) deal with information presented in visual diagrams (\cite{ARISTO_allen_2019}).  
    
    The line of research proposed here would concentrate specifically on the most visible difference between the way that humans seem to ``understand'' and the way in which machines do. Humans, in both everyday and scientific problems, reduce complex realities to set of powerful and concise rules.  In the scientific realm, the rules are often mathematical. In the more human realm they may be folkloric, as in ``a stitch in time saves nine,'' or ``there is more than one way to skin a cat.'' Skilled craftspeople know many such rules, and solve novel problems every day, by rethinking what they know, and formulating a useful (if temporary) rule for the situation at hand.


\bibliographystyle{plainnat}
\bibliography{RuleGame}

\begin{thebibliography}{33}
\providecommand{\natexlab}[1]{#1}
\providecommand{\url}[1]{\texttt{#1}}
\expandafter\ifx\csname urlstyle\endcsname\relax
  \providecommand{\doi}[1]{doi: #1}\else
  \providecommand{\doi}{doi: \begingroup \urlstyle{rm}\Url}\fi

\bibitem[Athalye and Sutskever(2017)]{turtle_example}
Anish Athalye and Ilya Sutskever.
\newblock Synthesizing robust adversarial examples.
\newblock \emph{arXiv preprint arXiv:1707.07397}, 2017.

\bibitem[Bongard(1967)]{bongard_pattern_1967}
M.~M. Bongard.
\newblock \emph{Pattern {{Recognition}}}.
\newblock {Hayden Book Co., Spartan Books.}, {Rochelle Park, NJ.}, 1967.
\newblock ISBN 978-0-8104-9165-6.

\bibitem[Boyle(2019)]{ARISTO_allen_2019}
Alan Boyle.
\newblock Allen {Institute}’s {Aristo} {AI} system finally passes an
  eighth-grade science test, September 2019.
\newblock URL
  \url{https://www.geekwire.com/2019/allen-institutes-aristo-ai-program-finally-passes-8th-grade-science-test/}.

\bibitem[Brynjolfsson et~al.(2018)Brynjolfsson, Mitchell, and
  Rock]{suitability_machine_learning}
Erik Brynjolfsson, Tom Mitchell, and Daniel Rock.
\newblock What can machines learn, and what does it mean for occupations and
  the economy?
\newblock \emph{AEA Papers and Proceedings}, 108:\penalty0 43--47, 2018.

\bibitem[Carter(2003)]{carter_2003}
W.~Craig Carter.
\newblock Ordinary differential equations from physical models, 2003.
\newblock URL
  \url{http://pruffle.mit.edu/3.016-2005/Lecture_20_web/node1.html}.

\bibitem[Fung(1989)]{Fung_1989}
Francis Cheong~Yiu Fung.
\newblock Framework for building rule-based machine diagnostic expert systems.
\newblock \emph{Knowledge-Based Systems}, 2\penalty0 (4):\penalty0 228--238,
  1989.

\bibitem[Galison et~al.(1997)]{galison1997image}
Peter Galison et~al.
\newblock \emph{Image and logic: A material culture of microphysics}.
\newblock University of Chicago Press, 1997.

\bibitem[García-Martín et~al.(2019)García-Martín, Rodrigues, Riley, and
  Grahn]{Energy_garcia-martin_estimation_2019}
Eva García-Martín, Crefeda~Faviola Rodrigues, Graham Riley, and Håkan Grahn.
\newblock Estimation of energy consumption in machine learning.
\newblock \emph{Journal of Parallel and Distributed Computing}, 134:\penalty0
  75--88, December 2019.
\newblock ISSN 0743-7315.
\newblock \doi{10.1016/j.jpdc.2019.07.007}.
\newblock URL
  \url{http://www.sciencedirect.com/science/article/pii/S0743731518308773}.

\bibitem[Goodfellow et~al.(2017)Goodfellow, Papernot, Huang, Duan, Abbeel, and
  Clark]{goodfellow_attacking_2017}
Ian Goodfellow, Nicolas Papernot, Sandy Huang, Rocky Duan, Pieter Abbeel, and
  Jack Clark.
\newblock Attacking {Machine} {Learning} with {Adversarial} {Examples},
  February 2017.
\newblock URL \url{https://openai.com/blog/adversarial-example-research/}.

\bibitem[Krakovna(2018)]{krakovna_specification_2018}
Victoria Krakovna.
\newblock Specification gaming examples in {AI} - master list : {Sheet}1, April
  2018.
\newblock URL \url{https://bit.ly/2skJE9C}.

\bibitem[Kuhn(1962)]{kuhn1962structure}
Thomas~S Kuhn.
\newblock \emph{The structure of scientific revolutions}.
\newblock University of Chicago Press, 1962.

\bibitem[Kuhn(1970)]{kuhn_structure_1970}
Thomas~S. Kuhn.
\newblock The {Structure} of {Scientific} {Revolutions}: {Second} {Edition},
  {Enlarged}.
\newblock In Otto Neurath, editor, \emph{International {Encyclopedia} of
  {Unified} {Science}}, volume IIn2, pages xii+174. University of Chicago
  Press, 2nd, enlarged edition, 1970.
\newblock URL
  \url{https://www.nemenmanlab.org/~ilya/images/c/c5/Kuhn-1970.pdf}.

\bibitem[Lenat(1995)]{lenat1995cyc}
Douglas~B Lenat.
\newblock Cyc: A large-scale investment in knowledge infrastructure.
\newblock \emph{Communications of the ACM}, 38\penalty0 (11):\penalty0 33--38,
  1995.

\bibitem[Levin et~al.(2019)Levin, Meng, Singh, and Zhu]{levin_jerry}
Owen Levin, Zihang Meng, Vikas Singh, and Xiaojin Zhu.
\newblock Fooling computer vision into inferring the wrong body mass index.
\newblock \emph{arXiv}, 2019.
\newblock URL \url{https://arxiv.org/pdf/1905.06916.pdf}.

\bibitem[Lim et~al.(1993)Lim, Walkup, and Vannier]{lim_et_al}
I~Lim, R.K. Walkup, and M.W. Vannier.
\newblock Rule based artificial intelligence expert system for determination of
  upper extremity impairment rating.
\newblock \emph{Computer methods and programs in bio-medicine}, 39\penalty0
  (3-4):\penalty0 203--211, 1993.

\bibitem[Linhares(2000)]{linhares_glimpse_2000}
Alexandre Linhares.
\newblock A glimpse at the metaphysics of {{Bongard}} problems.
\newblock \emph{Artificial Intelligence}, 121\penalty0 (1):\penalty0 251--270,
  2000.
\newblock URL
  \url{http://www.sciencedirect.com/science/article/pii/S0004370200000424}.

\bibitem[Lupyan(2013)]{lupyan_difficulties_2013}
G.~Lupyan.
\newblock The difficulties of executing simple algorithms: {{Why}} brains make
  mistakes computers don't.
\newblock \emph{Cognition}, 129\penalty0 (3):\penalty0 615--636, December 2013.
\newblock ISSN 0010-0277.
\newblock \doi{10.1016/j.cognition.2013.08.015}.
\newblock URL
  \url{http://www.sciencedirect.com/science/article/pii/S0010027713001728}.

\bibitem[Lupyan(2015)]{lupyan_paradox_2015}
G.~Lupyan.
\newblock The paradox of the universal triangle: Concepts, language, and
  prototypes.
\newblock \emph{Quarterly Journal of Experimental Psychology}, 2015.
\newblock \doi{10.1080/17470218.2015.1130730}.

\bibitem[Lupyan and Clark(2015)]{lupyan_words_2015}
G.~Lupyan and A.~Clark.
\newblock Words and the {{World}}: {{Predictive}} coding and the
  language-perception-cognition interface.
\newblock \emph{Current Directions in Psychological Science}, 24\penalty0
  (4):\penalty0 279--284, 2015.
\newblock \doi{10.1177/0963721415570732}.

\bibitem[Majid et~al.(2018)Majid, Roberts, Cilissen, Emmorey, Nicodemus,
  O'Grady, Woll, LeLan, de~Sousa, Cansler, Shayan, de~Vos, Senft, Enfield,
  Razak, Fedden, Tufvesson, Dingemanse, Ozturk, Brown, Hill, Guen, Hirtzel, van
  Gijn, Sicoli, and Levinson]{majid_differential_2018}
A.~Majid, Se{\'a}n~G. Roberts, Ludy Cilissen, Karen Emmorey, Brenda Nicodemus,
  Lucinda O'Grady, Bencie Woll, Barbara LeLan, Hil{\'a}rio de~Sousa, Brian~L.
  Cansler, Shakila Shayan, Connie de~Vos, Gunter Senft, N.~J. Enfield,
  Rogayah~A. Razak, Sebastian Fedden, Sylvia Tufvesson, Mark Dingemanse, Ozge
  Ozturk, Penelope Brown, Clair Hill, Olivier~Le Guen, Vincent Hirtzel, Rik van
  Gijn, Mark~A. Sicoli, and Stephen~C. Levinson.
\newblock Differential coding of perception in the world's languages.
\newblock \emph{Proceedings of the National Academy of Sciences}, 115\penalty0
  (45):\penalty0 11369--11376, November 2018.
\newblock ISSN 0027-8424, 1091-6490.
\newblock \doi{10.1073/pnas.1720419115}.
\newblock URL \url{http://www.pnas.org/content/115/45/11369}.

\bibitem[Mnih et~al.(2015)Mnih, Kavukcuoglu, Silver, Rusu, Veness, Bellemare,
  Graves, Riedmiller, Fidjeland, Ostrovski, Petersen, Beattie, Sadik,
  Antonoglou, King, Kumaran, Wierstra, Legg, and Hassabis]{Deep_Mind}
Volodymyr Mnih, Koray Kavukcuoglu, David Silver, Andrei~A. Rusu, Joel Veness,
  Marc~G. Bellemare, Alex Graves, Martin Riedmiller, Andreas~K. Fidjeland,
  Georg Ostrovski, Stig Petersen, Charles Beattie, Amir Sadik, Ioannis
  Antonoglou, Helen King, Dharshan Kumaran, Daan Wierstra, Shane Legg, and
  Demis Hassabis.
\newblock Human-level control through deep reinforcement learning.
\newblock \emph{Nature}, 518:\penalty0 229--533, 2015.

\bibitem[Molnar(2018)]{molnar}
Christoph Molnar.
\newblock \emph{Interpretable Machine Learning: A Guide for Making Black Box
  Models Explainable}.
\newblock 2018.
\newblock URL \url{https://christophm.github.io/interpretable-ml-book/}.

\bibitem[NYU(no date)]{noauthor_quotations_nodate}
NYU.
\newblock Quotations about {Archimedes}' {Lever}, no date.
\newblock URL
  \url{https://www.math.nyu.edu/~crorres/Archimedes/Lever/LeverQuotes.html}.

\bibitem[Papernot and Frosst(2019)]{papernot_how_2019}
Nicolas Papernot and Nicholas Frosst.
\newblock How to know when machine learning does not know, May 2019.
\newblock URL \url{http://cleverhans.io/security/2019/05/20/dknn.html}.

\bibitem[Patel(2017)]{Patel_Daily_Beast}
Neel~V. Patel.
\newblock Why doctors aren’t afraid of better, more efficient ai diagnosing
  cancer.
\newblock \emph{Daily Beast}, 2017.
\newblock URL
  \url{https://www.thedailybeast.com/why-doctors-arent-afraid-of-better-more-efficient-ai-diagnosing-cancer}.

\bibitem[Polson and Scott(2018)]{polson_aiq:_2018}
Nick Polson and James Scott.
\newblock \emph{{AIQ}: {How} {People} and {Machines} {Are} {Smarter}
  {Together}}.
\newblock St. Martin's Publishing Group, May 2018.
\newblock ISBN 978-1-250-18215-9.
\newblock Google-Books-ID: 35NUDwAAQBAJ.

\bibitem[Quinn and Eimas(1996)]{quinn_perceptual_1996}
P.C. Quinn and Peter~D. Eimas.
\newblock Perceptual {{Cues That Permit Categorical Differentiation}} of
  {{Animal Species}} by {{Infants}}.
\newblock \emph{Journal of Experimental Child Psychology}, 63\penalty0
  (1):\penalty0 189--211, October 1996.
\newblock \doi{10.1006/jecp.1996.0047}.
\newblock URL
  \url{http://www.sciencedirect.com/science/article/B6WJ9-45MGWJP-V/2/7f074767a379292bd058aee27d94009d}.

\bibitem[Rissanen(1989)]{rissanen1989stochastic}
Jorma Rissanen.
\newblock \emph{Stochastic complexity in statistical inquiry}.
\newblock World Scientific, 1989.

\bibitem[Rogers and McClelland(2004)]{rogers_semantic_2004}
T.T. Rogers and J.L. McClelland.
\newblock \emph{Semantic Cognition: {{A}} Parallel Distributed Processing
  Approach}.
\newblock {Bradford Book}, {Cambridge, MA}, 2004.

\bibitem[Rumelhart et~al.(1986)Rumelhart, McClelland, and {the PDP Research
  Group}]{rumelhart_parallel_1986}
D.E. Rumelhart, J.L. McClelland, and {the PDP Research Group}.
\newblock \emph{Parallel {{Distributed Processing}}: {{Explorations}} in the
  {{Microstructure}} of {{Cognition}}, {{Volumes}} 1 and 2}.
\newblock {MIT Press}, {Cambridge, MA}, 1986.

\bibitem[Shortliffe et~al.(1975)Shortliffe, Davis, Axline, Buchanan, Green, and
  Cohen]{shortliffe1975computer}
Edward~H Shortliffe, Randall Davis, Stanton~G Axline, Bruce~G Buchanan,
  C~Cordell Green, and Stanley~N Cohen.
\newblock Computer-based consultations in clinical therapeutics: explanation
  and rule acquisition capabilities of the mycin system.
\newblock \emph{Computers and biomedical research}, 8\penalty0 (4):\penalty0
  303--320, 1975.

\bibitem[Silver et~al.(2017)Silver, Schrittwieser, Simonyan, Antonoglou, Huang,
  Guez, Hubert, Baker, Lai, Bolton, et~al.]{silver2017mastering}
David Silver, Julian Schrittwieser, Karen Simonyan, Ioannis Antonoglou, Aja
  Huang, Arthur Guez, Thomas Hubert, Lucas Baker, Matthew Lai, Adrian Bolton,
  et~al.
\newblock Mastering the game of go without human knowledge.
\newblock \emph{Nature}, 550\penalty0 (7676):\penalty0 354, 2017.

\bibitem[{von Neumann}(2012)]{von_neumann_computer_2012}
John {von Neumann}.
\newblock \emph{The {{Computer}} and the {{Brain}}}.
\newblock {Yale University Press}, {New Haven, Conn. ; London}, 3 edition
  edition, August 2012.
\newblock ISBN 978-0-300-18111-1.

\end{thebibliography}
\end{document}